\documentclass[letterpaper, 10 pt, conference]{ieeeconf}  

\IEEEoverridecommandlockouts                              

\usepackage{graphics} 
\usepackage{epsfig} 
\usepackage{times} 
\usepackage{amsmath} 
\usepackage{amssymb}  
\usepackage{hyperref}
\usepackage{tabularx}
\usepackage{comment}
\usepackage{algorithm}  
\usepackage{algorithmic}
\usepackage{subfigure}
\usepackage{booktabs}
\usepackage{array}
\usepackage{multirow}
\usepackage{stfloats}
\usepackage{longtable}
\usepackage{rotating}
\title{\LARGE \bf
    Critic PI2: Master Continuous Planning via Policy Improvement with Path Integrals and Deep Actor-Critic Reinforcement Learning
}

\author{Jiajun Fan$^{1}$, He Ba$^{1}$, Xian Guo$^{1}$, Jianye Hao$^{2}$ 
\thanks{*Jiajun Fan and He Ba contributed equally to this work.}
\thanks{$^{1}$Jiajun Fan, He Ba, Xian Guo are with the College of Artificial Intelligence, Nankai University, Tianjin 300350, China. {\tt\small corresponding author at: guoxian@nankai.edu.cn}
    }%
\thanks{$^{2}$ Jianye Hao is with Huawei Noah’s Ark Lab, Huawei Technologies, China
       }%
}

\def\BibTeX{{\rm B\kern-.05em{\sc i\kern-.025em b}\kern-.08em
    T\kern-.1667em\lower.7ex\hbox{E}\kern-.125emX}}
\begin{document}

\maketitle
\thispagestyle{empty}
\pagestyle{empty}

\begin{abstract}
Constructing agents with planning capabilities has long been one of the main challenges in the pursuit of artificial intelligence. Tree-based planning methods from AlphaGo to Muzero have enjoyed huge success in discrete domains, such as chess and Go. Unfortunately, in real-world applications like robot control and inverted pendulum, whose action space is normally continuous, those tree-based planning techniques will be struggling. To address those limitations, in this paper, we present a novel model-based reinforcement learning frameworks called Critic PI2, which combines the benefits from trajectory optimization, deep actor-critic learning, and model-based reinforcement learning. Our method is evaluated for inverted pendulum models with applicability to many continuous control systems. Extensive experiments demonstrate that Critic PI2 achieved a new state of the art in a range of challenging continuous domains. Furthermore, we show that planning with a critic significantly increases the sample efficiency and real-time performance. Our work opens a new direction toward learning the components of a model-based planning system and how to use them.
\end{abstract}

\begin{figure*}[!t]
	\centering
	\subfigure{
		\label{fig_1_a}
		\centering
		\includegraphics[width=0.4\textwidth,height=0.25\textheight]{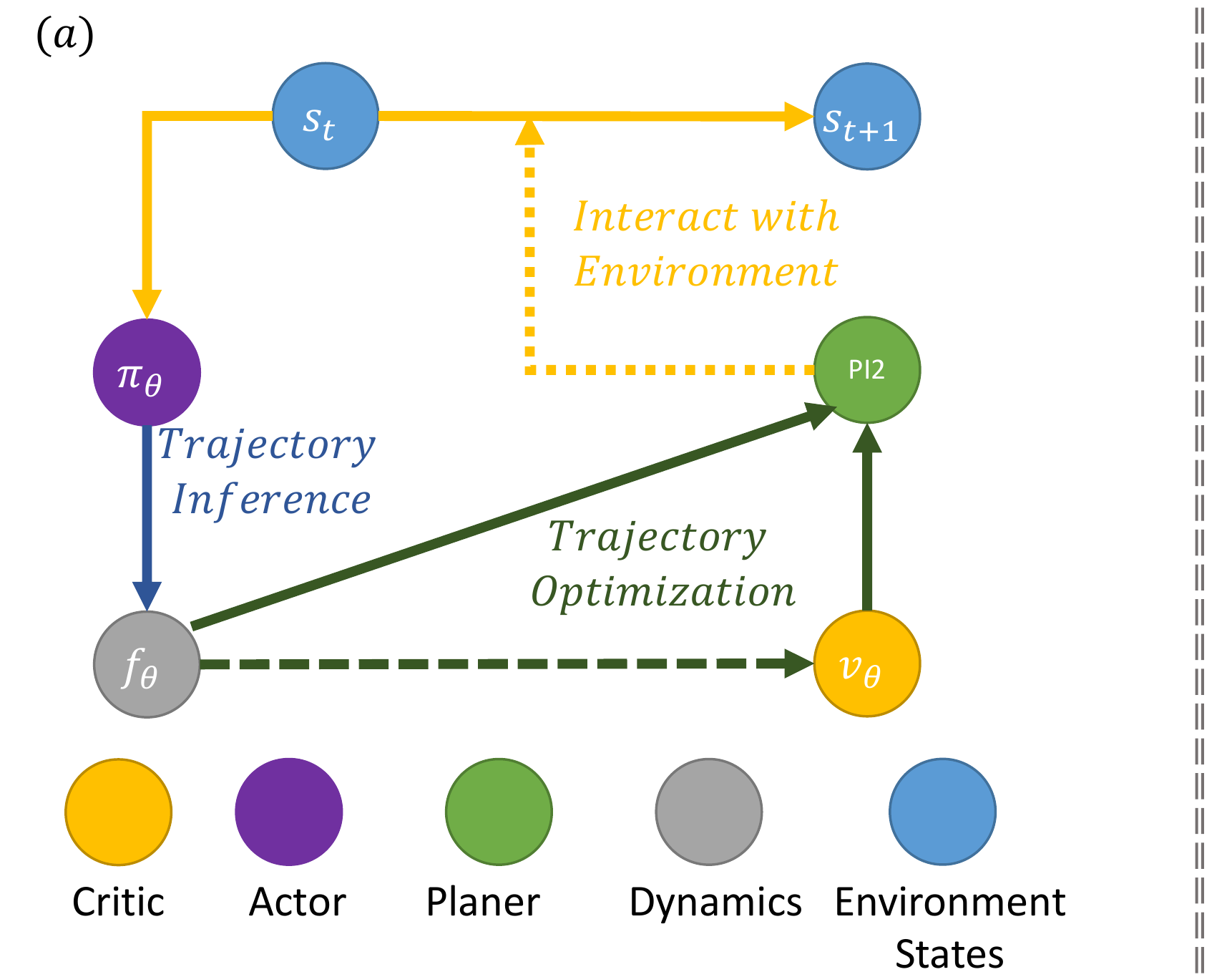}
	}%
	\subfigure{
		\label{fig_1_b}
		\centering
		\includegraphics[width=0.5\textwidth,height=0.25\textheight]{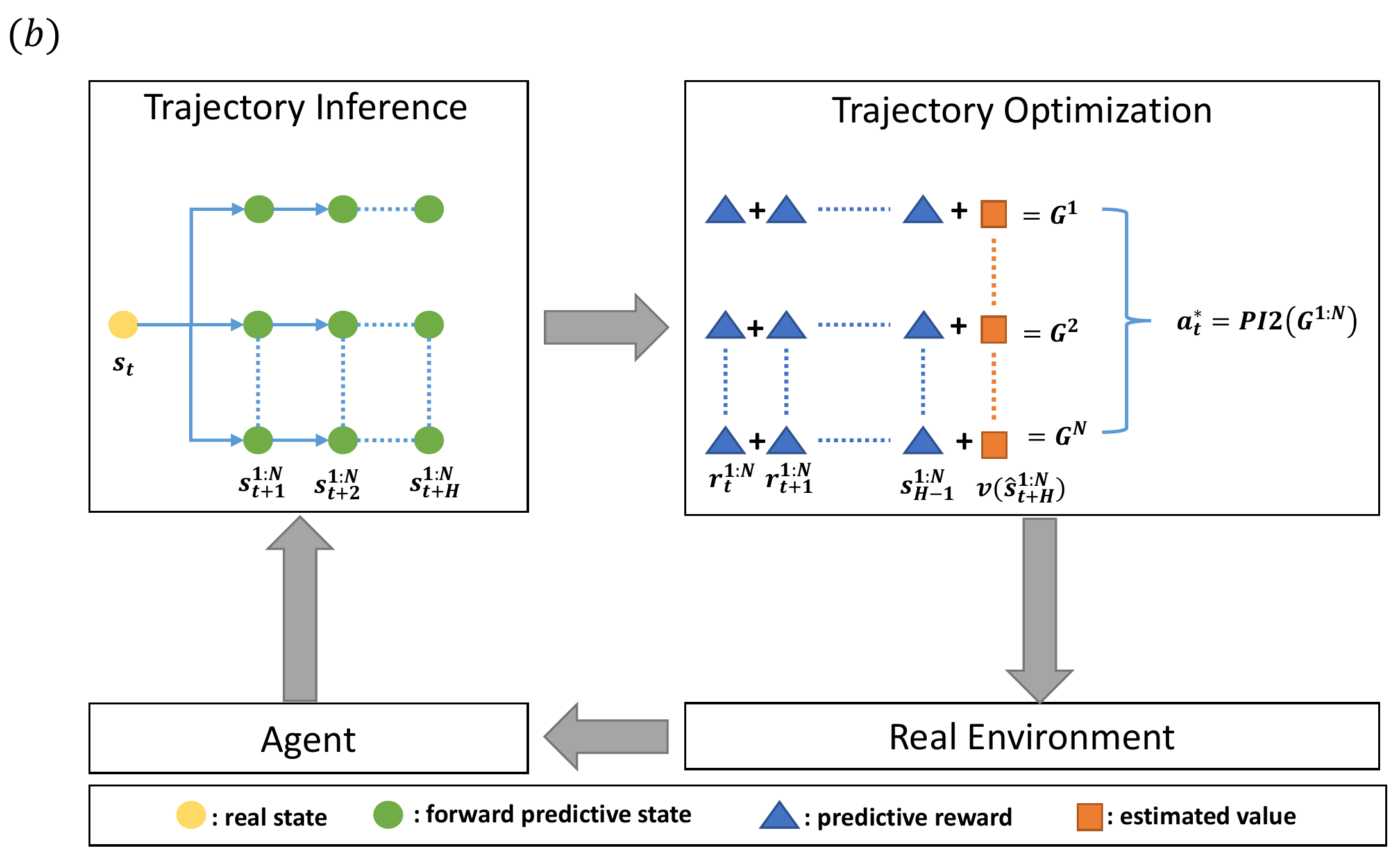}
	}%
	\centering
	\caption{Overview of the Critic PI2 algorithm. \textbf{Fig. \ref{fig_1_a}} illustrates the general frameworks of Critic PI2. Specially, at each time step agent will repeat the cycle of trajectory inference and trajectory optimization, thus obtaining an expert action to interact with the environment. \textbf{Fig. \ref{fig_1_b}} shows more details about the trajectory inference and trajectory optimization. Specially, when interacting with the environment (e.g. a robot system), the agent uses the predictive dynamics $\hat{f_{\theta}}$ to predict the future states over a finite horizon $\mathcal{H}$ with an actor and thus generate $\mathcal{N}$ trajectory, which is called trajectory inference. Then an expert action $a^{*}$ is obtained after 
$\mathcal{K}$ times trajectory optimization from the planner  with the assistance of a critic using Critic PI2 algorithm. After stepping this expert action in the environment, agent will draw the next observation, and then repeat in the next time step.}
	\label{fig_1}
\end{figure*}

\section{INTRODUCTION}


Deep reinforcement learning (DRL) methods have shown recent success on continuous control tasks in robotics systems in simulation  \cite{cbto1} \cite{cbto2} \cite{cbto3}. Reinforcement learning (RL) algorithms generally fall into one of two categories: model-free techniques, which learn a direct mapping from states to actions, and model-based approaches, which build a predictive model of an environment and derive a controller from it. Model-free methods have shown promise as a general-purpose tool for learning complex policies from raw state inputs   \cite{c1} \cite{c2} \cite{c3}, and such methods are applied using no prior knowledge of the systems, leading to problematic sample complexity and thus long training times.  However, when dealing with real-world physical systems, for which data collection can be an arduous process, model-free methods become struggling due to the sample complexity and model-based approaches are appealing due to their comparatively fast learning. Unfortunately, model accuracy acts as a bottleneck to policy quality, often causing model-based approaches to perform worse asymptotically than their model-free counterparts. What's worse, little can be said about the stability or robustness of these resulting control policies, even if more traditional model-based optimal control solutions exist for these same systems.

Different from traditional RL, path-integral-based RL, such as policy improvement with path integrals (PI2), within the framework of stochastic optimal control, requires far fewer iterations and guarantees optimality and reliable training convergence  \cite{tnnis25} \cite{tnnis26} \cite{tnnis27}. As a matter of fact, when training the traditional RL using approximated dynamics, instead of the authentic one of the studied practical system, which thus causing an approximation error. Unfortunately, this inaccuracy often degrades the performance of the system and thus needs to be carefully addressed. Inspired by this fact, the model-based Lyapunov method, which can ensure the robustness and stability of a nonlinear system even in the presence of various uncertainties, is introduced into the original model-based RL to solve a range of control problems. A similar model-based RL approach that can notably increase the learning speed is proposed in  \cite{tnnis28}; however, its application scope is limited to video games.

In this paper, we investigate how to most effectively combine the superiority of  deep actor-critic learning, policy optimization and model-based reinforcement learning in continuous domains. Specifically, we first introduce the Path-Integrals-based trajectory optimization algorithm into model-based reinforcement learning, thus obtaining the Deep PI2 (DPI2) algorithm, which enables agents the ability to plan in continuous space. However, the model predictive error caused by model approximation using deep neural networks hinders the further improvement of the performance. To outbreak this limitation, we draw insights from model-free reinforcement learning to introduce the Deep Actor-Critic Learning into our DPI2 frameworks, thus obtaining the Critic PI2 algorithm  (summarized in \textbf{Fig. \ref{fig_1}}), which introduces a critic to assist the planner and thus reduce the influence of model approximation error.  

Our core contribution is a practical framework called Critic PI2 built on these insights, which can ease the burden of the predictive dynamic model by one-step inference and simultaneously improve the effectiveness of the planning algorithm by using a novel path-integrals-based trajectory optimization algorithm. The experiment results can support the conclusion that using a critic to assist the predictive model can effectively ease the misleading of incorrect predictive models, thus guaranteeing a good performance.

The remaining of the paper is organized as follows. In Section \ref{phaze_RELATED_WORK}, we introduce some prior work in related domains. In Section \ref{phase_preliminaries}, some preliminaries of our algorithm will be introduced. More details on the whole framework will be proposed in Section \ref{phase_CRITIC_PI2_FRAMEWORKS}, and adequate simulation experiments, reasonable analysis and credible results of the proposed algorithm will be demonstrated in Section \ref{phase_EXPERIMENTS}. Finally, Section \ref{phase_conclusion} concludes.

\section{PRIOR WORK}
\label{phaze_RELATED_WORK}

Model-free reinforcement learning algorithms based on Qlearning  \cite{mpc13} \cite{mpc2} \cite{mpc9}, actor-critic methods  \cite{mpc14} \cite{mpc15} \cite{mpc16} and policy gradients  \cite{mpc3} \cite{mpc17} like Deep Deterministic Policy Gradient (DDPG) have been shown to learn very complex skills in high-dimensional state spaces, including simulated robotic locomotion, driving, video game playing, and navigation. 
However, the high sample complexity of purely model-free algorithms has made them difficult to be implemented in the real world, where sample collection is limited by the constraints of real-time operation. 

Model-based algorithms are known in general to outperform model-free learners in terms of sample complexity  \cite{mpc4}, and in practice have been applied successfully to control robotic systems both in simulation and in the real world, such as pendulums  \cite{mpc6}, legged robots  \cite{mpc18}, swimmers  \cite{mpc19}, and manipulators  \cite{mpc20}. However, the most efficient model-based algorithms have used relatively simple function approximators, such as Gaussian processes  \cite{mpc21} \cite{mpc22}, time-varying linear models  \cite{mpc23} \cite{mpc24}, and mixtures of Gaussians  \cite{mpc25}, and the model error caused by the approximation tends to cripple the performance of model-based approaches, which is also known as model-bias. As is discovered previously, even a small model error can severely degrade multi-step rollouts since the model error will compound as the steps increases. Hence the predicted states will move out of the region where the model has high accuracy after a few steps. 

    Several previous works have been proposed to alleviate the influence of compounding model error in different ways. For example, some work  \cite{mpc} hybrids model-based and model-free algorithm by initializing a model-free learner with a model-based algorithm like model predictive control(MPC). However, the hybrid algorithm can only achieve better final performance at the expense of much higher sample complexity. Although the model-based part of this kind of method  \cite{mpc} is far more  efficient in sampling and more flexible than task-specific policies learned with model-free reinforcement learning, their asymptotic performance is usually worse than model-free learners due to the influence of model bias.
\begin{figure*}[!t]
	\centering
	\subfigure{
		\centering
		\includegraphics[width=0.3\textwidth,height=0.2\textheight]{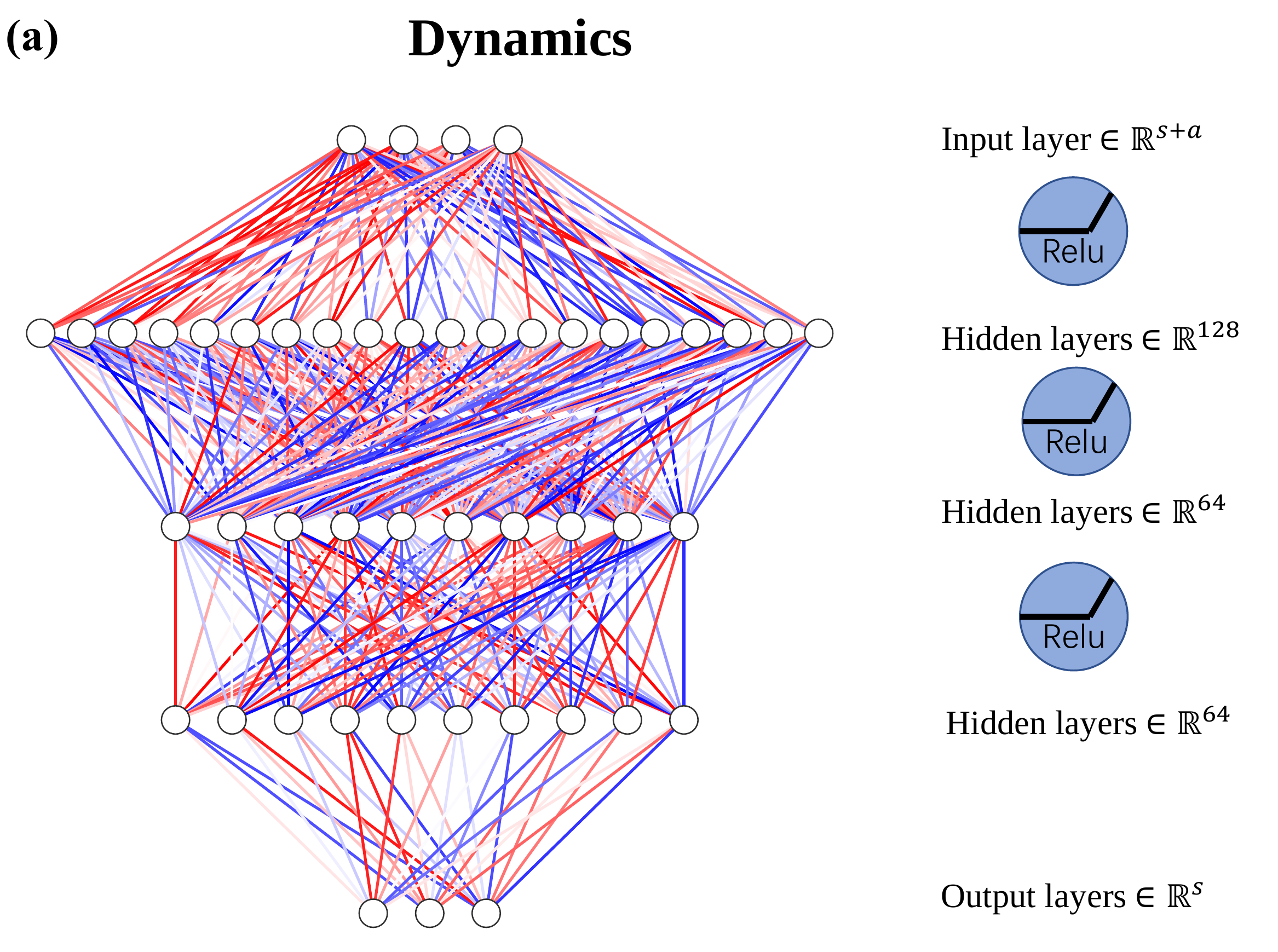}
        \label{fig_dynamics}
	}%
	\subfigure{
		\centering
		\includegraphics[width=0.3\textwidth,height=0.2\textheight]{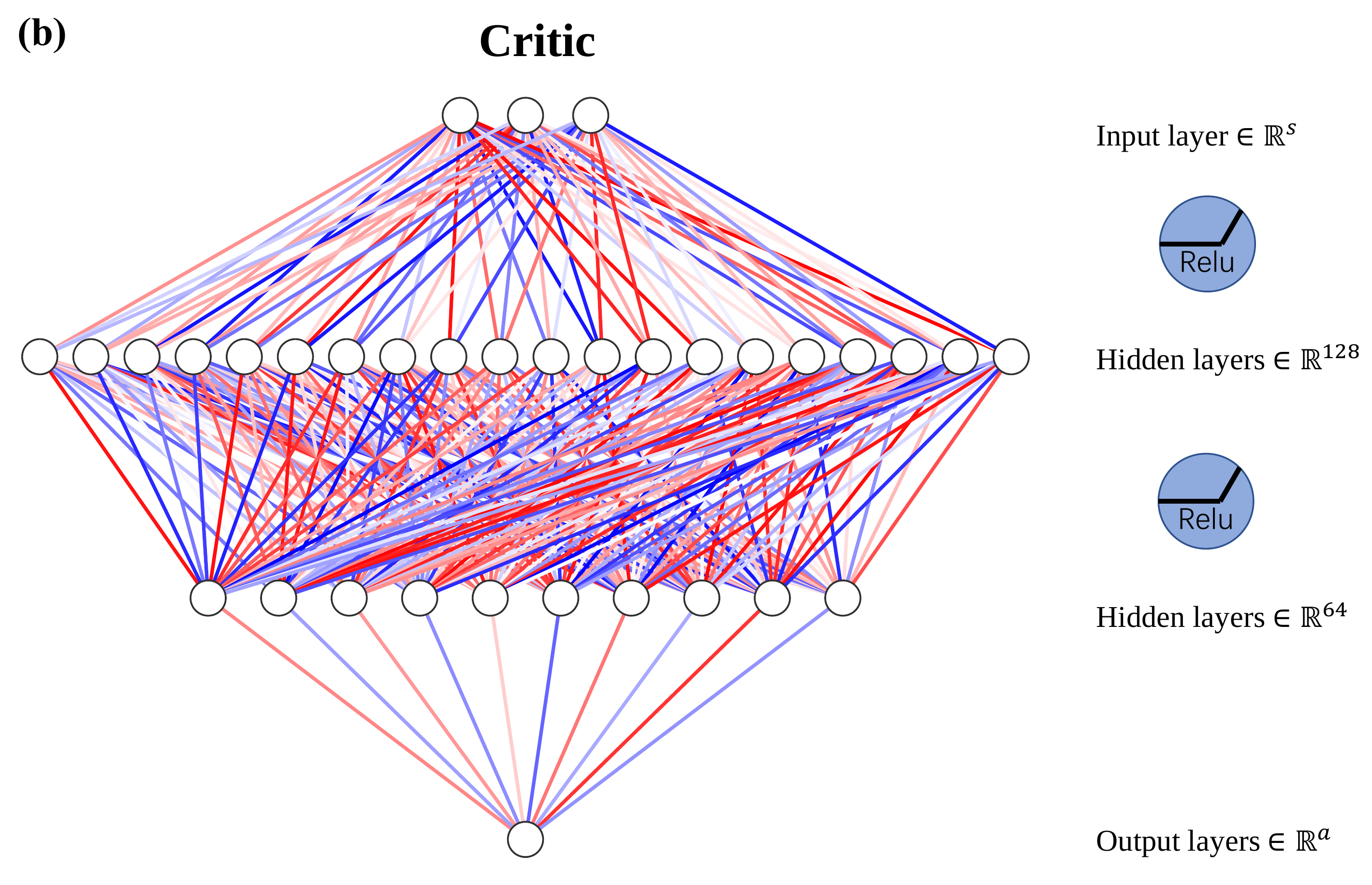}
		\label{fig_critic}
	}%
	\subfigure{
		\centering
		\includegraphics[width=0.3\textwidth,height=0.2\textheight]{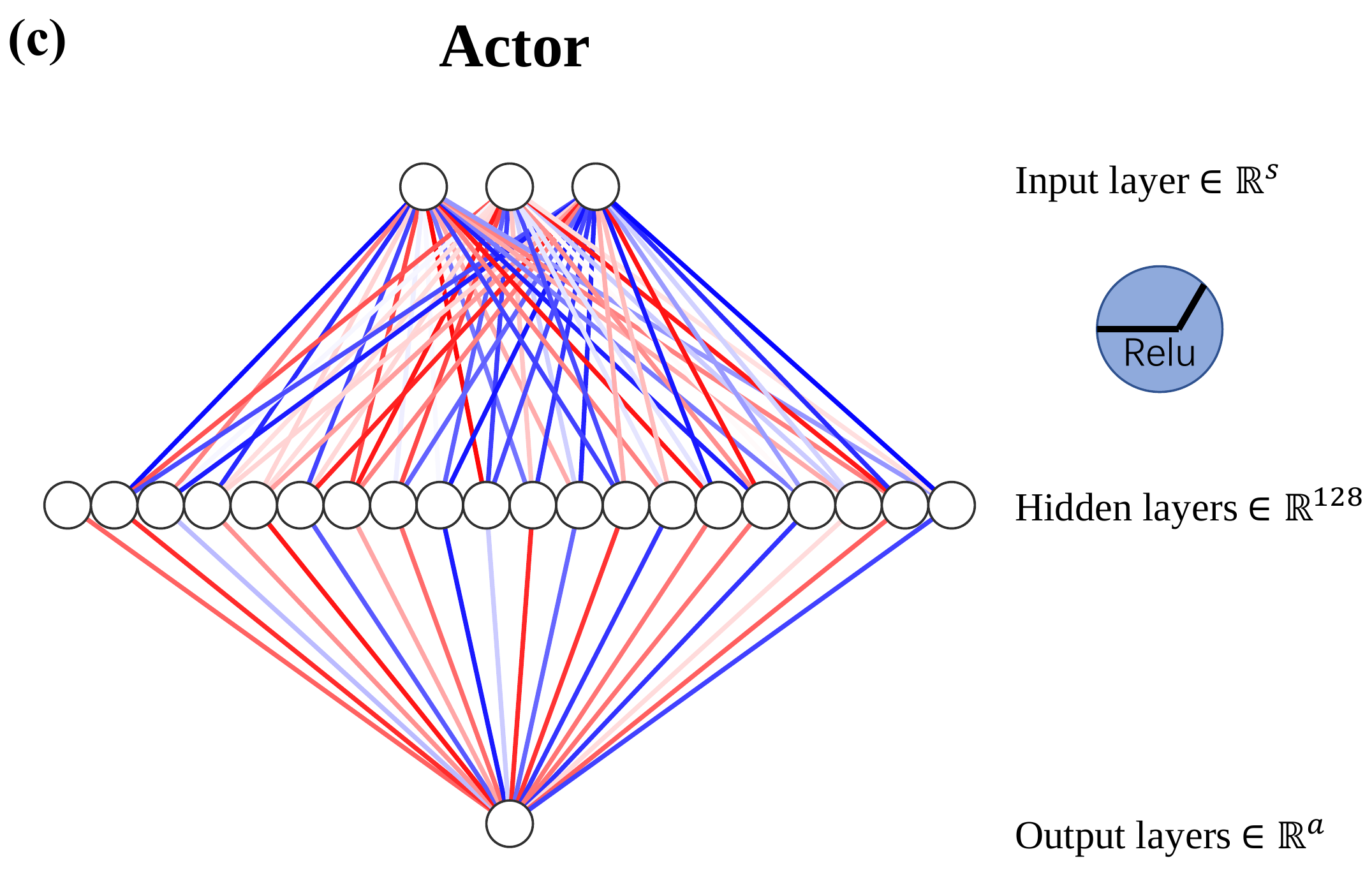}
		\label{fig_actor}
	}%
	\centering
	\caption{Network architecture of the core component of Critic PI2 (Actor, Critic, Dynamics as \textbf{Fig. \ref{fig_1}}). Among them, actor approximates the optimal policy which maps the system state $s_t$ to an optimal action $u_t^*$, critic represent the state value function trained using v-trace and dynamics is trained to construct the model dynamics of the environment. All three Network is trained independently and centrally.} 
	\label{fig_Network}
\end{figure*}

\section{PRELIMINARIES}
\label{phase_preliminaries}
\subsection{Reinforcement Learning}
We consider a Markov decision process (MDP), defined by the tuple  $\left(\mathcal{S}, \mathcal{A}, p, r, \gamma, \rho_{0}\right)$. $\mathcal{S}$ and $\mathcal{A}$ are the state and action spaces, respectively, and $\gamma \in(0,1)$ is the discount factor. The dynamics or transition distribution are denoted as $p\left(s^{\prime} \mid s, a\right)$, the initial state distribution as $\rho_0(s)$, and the reward function as $r(s,a)$. The goal of reinforcement learning is to find the optimal policy $\pi^*$ that maximizes the expected sum of discounted rewards, denoted by $\eta$:
\begin{equation}
\label{eq_accmulate_reward}
\pi^{*}=\underset{\pi}{\operatorname{argmax}} \eta_{\pi}(\tau) = \underset{\pi}{\operatorname{argmax}} E_{\pi}\left[\sum_{t=0}^{\infty} \gamma^{t} r\left({s}_{t}, a_{t}\right)\right]
\end{equation}

The dynamics $p\left(s^{\prime} \mid s, a\right)
$ are assumed to be unknown. Model-based reinforcement learning methods
aim to construct a model of the transition distribution, $f_{\theta}\left(s^{\prime} \mid s, a\right)
$, with the MDP data collected from interaction, typically using imitation learning.

\subsection{Stochastic Optimal Control}

The general stochastic dynamical system is expressed as follows

\begin{equation}
\label{Stochastic Dynamical System}
\begin{aligned}
\boldsymbol{\dot{x}_{t}} &=\boldsymbol{f}\left(\boldsymbol{{x}_{t}}, t\right)+\boldsymbol{G}\left(\boldsymbol{{x}_{t}}\right)\left(\boldsymbol{u_{t}}+\boldsymbol{\varepsilon_{t}}\right) \\
&=\boldsymbol{f_{t}}+\boldsymbol{G_{t}}\left(\boldsymbol{u_{t}}+\boldsymbol{\varepsilon_{t}}\right)
\end{aligned}
\end{equation}
where $\boldsymbol{u}_{t} \in \mathfrak{R}^{p \times 1}$ represents the control vector, $\boldsymbol{f}_{t} \in \mathfrak{R}^{n \times 1}$ stands for the system dynamics, $\boldsymbol{G}_{t} \in \mathfrak{R}^{n \times p}$ donates the control matrix, $\boldsymbol{{x}_{t}} \in \mathfrak{R}^{n \times 1}$ is the system’s state, and $\boldsymbol{\varepsilon_{t}} \in \mathfrak{R}^{p \times 1}$ is the Gaussian noise submitting to $N (0, \Sigma_{\varepsilon})$. 

The performance criterion function for a path $\tau$ starting at time $t_i$ in state $x_t$ and ending at time $t_f$ is defined as follows  \cite{cac15}
\begin{equation}
\label{J_fuction}
J\left(\tau\right)=\phi_{t_{f}}+\int_{t_{i}}^{t_{f}} L\left[{x}_{t}, u_{t}, t\right] d t
\end{equation}
wherein $\phi_{t_{f}}$ is terminal value, and $\int_{t_{i}}^{t_{f}} L\left[{x}_{t}, u_{t}, t\right] d t$ stands for process value. 
The immediate cost is defined as follows
\begin{equation}
\label{immediate cost}
L_{t}=L\left[{x}_{t}, u_{t}, t\right]=q_{t}+\frac{1}{2} u_{t}^{T} \boldsymbol{R} u_{t}
\end{equation}
where $q_{t}=q\left(\boldsymbol{x}_{t}, t\right)$ is the cost function related to states, $\boldsymbol{R}$ represents the coefficient matrix.
The goal of stochastic optimal control is to seek out the control $u_t$ to minimize the following performance criterion function
\begin{equation}
\label{performance criterion function}
\min\quad V\left(\boldsymbol{x}_{t}\right)=V_{t}=\min _{t_{i}:t_{f}} E_{\tau}\left[J\left(\tau\right)\right] 
\end{equation}
where $E_{\tau}[\cdot]$ is the expectation of all trajectories' performance criterion function starting from $\boldsymbol{x}_{t}$.

\subsection{Trajectory Optimization With Path Integral}

 It is very difficult to obtain the analytical solution for optimization problem as \eqref{performance criterion function}, instead, with path integral algorithm, which is a numerical method used to solve stochastic optimal control problems, for which the goal is to minimize a performance criterion for a stochastic dynamical system  \cite{cac15}, we can propose its numerical solutions after iteration convergence of \eqref{update_u_equ}. 
\begin{equation}\label{update_u_equ}
u_{t}^{i}=\int P\left(\tau_{i}\right) u\left(\tau_{i}\right) d \tau_{i}
\end{equation}
where $P\left(\tau_{i}\right)$ is defined   \cite{cac16} as follows
\begin{equation}
\label{probility equ}
P\left(\tau_{i}\right)=\frac{e^{-\frac{1}{\lambda} S\left(\tau_{i}\right)}}{\int e^{-\frac{1}{\lambda} S\left(\tau_{i}\right)} d \tau_{i} }
\end{equation}

Equation \eqref{probility equ} can be approximated to \eqref{probility equ_approvimate} for the convenience of implementation.

\begin{equation}
\label{probility equ_approvimate}
P\left(\tau_{i}\right)=\frac{e^{-\frac{1}{\lambda} S\left(\tau_{i}\right)}}{\sum_{i=1}^{N} e^{-\frac{1}{\lambda} S\left(\tau_{i}\right)}}
\end{equation}
wherein $S\left(\tau_{i}\right)$ is a normalized version of the path cost defined as follows
\begin{equation}\label{normalized_path cost}
S\left(\tau_{i}\right)= \frac{C\left(\tau_{i}\right)-\min (\boldsymbol{C})}{\max (\boldsymbol{C})-\min (\boldsymbol{C})}
\end{equation}
where $C\left(\tau_{i}\right)$ represents the cost function similar to $J\left(\tau_{i}\right)$, $\boldsymbol{C}$ represents the cost function vector for $\mathcal{N}$ trajectories. 

It's obvious that we can unite the representations of reinforcement learning and stochastic optimal control by making the reward function equal  to the negative of the cost function. For ease of demonstration, we will use the representations of RL manners in the following sections.

\section{CRITIC PI2 FRAMEWORKS}
\label{phase_CRITIC_PI2_FRAMEWORKS}
 In this section, more implementation details of the Critic PI2 will be revealed. We introduce how to learn the model predictive network in \ref{Dynamics Network}, detail how to train actor-critic architecture in Sec \ref{Actor Critic Network}, and demonstrate the whole frameworks of Critic PI2 in Sec \ref{Critic PI2}.

\subsection{Dynamics Network}
\label{Dynamics Network}
Traditional model-based reinforcement learning algorithms always learn dynamic network via imitation learning, whose structure can be found in \textbf{Fig. \ref{fig_dynamics}}, which maps from current state $s_t$ and current action $a_t$ to next state $s_{t+1}$. Different from that approach, we choose to learn a differential target $s_{t+1}-s_t$ as \eqref{equ_dynamics}, which can reduce the influence of the environment noise and improve the learning effect.
\begin{equation}
\label{equ_dynamics}
s_{t+1}-s_{t} = \hat{f}_{\theta}(s_t,a_t)
\end{equation}

In the normal manner of PI2  \cite{2016path}, the discounted cumulative reward $R(\tau)$ is used to guide the improvement direction of the policy. However,  \cite{mbpo} proves the fact that even a small model error will  cause great approximation bias  of the value function obtained by simulating k steps in the incorrect predictive dynamics shown as \eqref{equ_mbpo}.
\begin{equation}
\label{equ_mbpo}
 \eta^{\mathrm{predictive}}[\pi] - \eta[\pi] \leq bias 
\end{equation}
\begin{equation*}
bias = 2*r_{\max }\left[\frac{\gamma^{k+1} \epsilon_{\pi}}{(1-\gamma)^{2}}+\frac{\gamma^{k}+2}{(1-\gamma)} \epsilon_{\pi}+\frac{k}{1-\gamma}\left(\epsilon_{m}+2 \epsilon_{\pi}\right)\right]    
\end{equation*}
$\epsilon_{m}= \max _{t} E_{s \sim \pi_{D, t}}\left[D_{T V}\left(p\left(s^{\prime}, r \mid s, a\right) \| p_{\theta}\left(s^{\prime}, r \mid s, a\right)\right)\right]$, which can be estimated in practice by measuring the validation loss of the model on the time-dependent state distribution of the data-collecting policy $\pi_D$, and the distribution shift by the maximum total-variation distance is remembered as $\epsilon_{\pi}$, which means
$\max _{s} D_{T V}\left(\pi \| \pi_{D}\right) \leq \epsilon_{\pi}$.

It reminds us of fact that the estimation error of the $R(\tau)$ is positive correlated with the length of trajectory inference. Thus, we can use the N step time difference (TD) methods to estimate the $R(\tau)$ as \eqref{eq_td} instead of  using  Monte Carlo estimation as \eqref{eq_mc} directly.

\begin{align}
    \label{eq_td}
     R^{(n)}(\tau) = \sum_{k=0}^n \gamma^{k}*r_{t+k+1} +\gamma^{n}*V_{\pi}(s_{t+n+1}) 
\end{align}
\begin{align}
    \label{eq_mc}
     R(\tau) = \sum_{t=0}^{\infty} \gamma^k r_{t+k+1}
\end{align}
 
To apply the N step TD method, we must construct an approximator to estimate the state value function $V_{\pi}(s)$, which will be proposed in the next Section.

\subsection{Actor Critic Network }
\label{Actor Critic Network}
Deep neural networks have been used to represent the value function $V_\pi(s)$ like \textbf{Fig. \ref{fig_critic}}. With the help of it, we can directly get the approximation of $\eta(\pi)$ by using the one-step TD methods as \eqref{eq_td}, which just need to infer only one step in the incorrect dynamics. After combining the superiority of model-free reinforcement learning and model-based reinforcement learning, even with extremely little sample the agent can still achieve a remarkable score in a range of continuous control tasks on the mujoco platform. To further improve the learning efficiency, we use the  V-trace target inspired 
by \cite{2018impala} as our value function's target, which can be computed as \eqref{v_trace}:
;\begin{equation}
\label{v_trace}
\begin{aligned}
    v_s = V(x_s) + \sum_{t=s}^{s+n-1}\gamma^{t-s}(\prod_{i=s}^{t-1}c_i)\delta_t V \\
    \delta_t V = \rho_t (r_t+\gamma V(s_{t+1}-V(s_t)))\\
    \rho_t = \min(\Bar{\rho},\frac{\pi(a_t|s_t)}{\mu(a_t|s_t)}) \\
    c_i = \min(\Bar{c},\frac{\pi(a_i|s_i)}{\mu(a_i|s_i)})
\end{aligned}
\end{equation}
where $\pi$ is the behavior policy and $\mu$ is the target policy.

The policy is constructed as a multi-variate Gaussian distribution approximated by a deep neural network, whose variance is normally fixed and the mean is predicted by deep neural networks described in \textbf{Fig. \ref{fig_actor}}, which is learned via imitation learning.
\subsection{Critic PI2}
\label{Critic PI2}

After showing how to learn the dynamics, actor and critic, there is still a core role in \textbf{Fig. \ref{fig_1_a}} waits to be revealed, which is called planner.

 Different from traditional PI2  \cite{2016path}, we apply a greedy strategy into the construction of the planner during optimization to further improve the optimality  of the planning algorithm. Specially, the previous PI2 method  will select K actions from the Gaussian  distribution and calculate their corresponding cumulative reward. Then it will calculate the PI2 optimal action $a_{opt}$ with \eqref{eq_mc}. After generating the optimal action, it will use the optimal action as the mean of the Gaussian Distribution to generate K actions and then repeat the previous procedures  several times. Finally, it will use the final optimal action as the expert action $a_t^{*}$. However, with the assistance of the critic network, our method can easily calculate a trajectory's reward with \eqref{eq_td}, which is much faster, more accurate, and has low variance. Hence, we propose a greedy strategy. Our agent will remember the best expert action. Extensive experiments prove the fact that the greedy  strategy is critical to performance.

After agents completely learn the four core components of Critic PI2 (Actor, Critic, Planner, Dynamics), we now can demonstrate our whole frameworks in \textbf{Fig \ref{fig_1}}. More implementation details can be found in \textbf{Algorithm \ref{alg:Critic PI2}}.

\begin{algorithm}[h]
	\caption{Critic Policy Improvement with Path Integrals (Critic PI2)}
	\label{alg:Critic PI2}
	\begin{algorithmic}[1]
		\STATE{Initialize policy network $\pi_{\phi}$, dynamic network $f_{\theta}$, value network $v_{\theta'}$, replay buffer $\mathcal{D}$ }
		\FOR{ $\mathcal{N}\ episodes$ }
		\STATE{Draw an observation $o_t$ from the environment}
		\FOR{ $\mathcal{T}\ time\_steps$ }
		\FOR{$\mathcal{M}\ Iterations$}
      \STATE{ Generate $\mathcal{K}$ trajectories from dynamics with $\mathcal{K}$ actors}
    \STATE {Obtain $R(\tau_i)$ using \eqref{eq_td}}
     \STATE {Generate an expert action from planner with the assistance of critic}
     \STATE {Actor learns from the expert}
        \ENDFOR
		\STATE {Obtain the expert action with greedy strategy}
		\STATE{Take the expert action in the environment, draw next observation $o_{t+1}$ and reward $r_{t}$}
		\STATE{Add experience to replay buffer}
		\ENDFOR
		\FOR{$\mathcal{E}\ Epochs$}
		\STATE{Update Actor, Dynamics, Critic centrally}
		\ENDFOR
		\ENDFOR
	\end{algorithmic}
	\hspace*{0.02in} {\bf Return:} $f_{\theta}^*,v_{\theta'}^*,\pi_{\phi}^*$
\end{algorithm}

\section{EXPERIMENTS}
\label{phase_EXPERIMENTS}
Our experiments aim to answer the following three questions:
\begin{enumerate}
    \item How does Critic PI2 perform compared with model-free RL methods and previous state-of-the-art model-based RL methods using model predictive control?
    \item Whether Critic PI2 obtain the optimality at the cost of real-time performance?
    \item What are the critical components of our overall algorithm?
\end{enumerate}

\begin{figure}[!t]
	\centering
	\subfigure{
		\centering
		\includegraphics[scale=0.7]{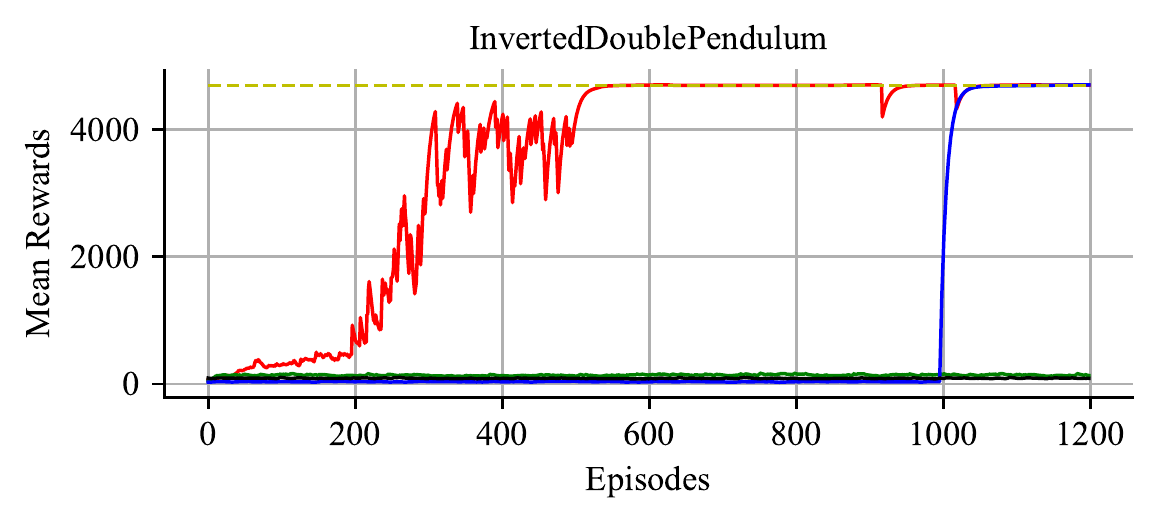}
	}%
\vspace{-5mm}
	\subfigure{
		\centering
		\includegraphics[scale=0.7]{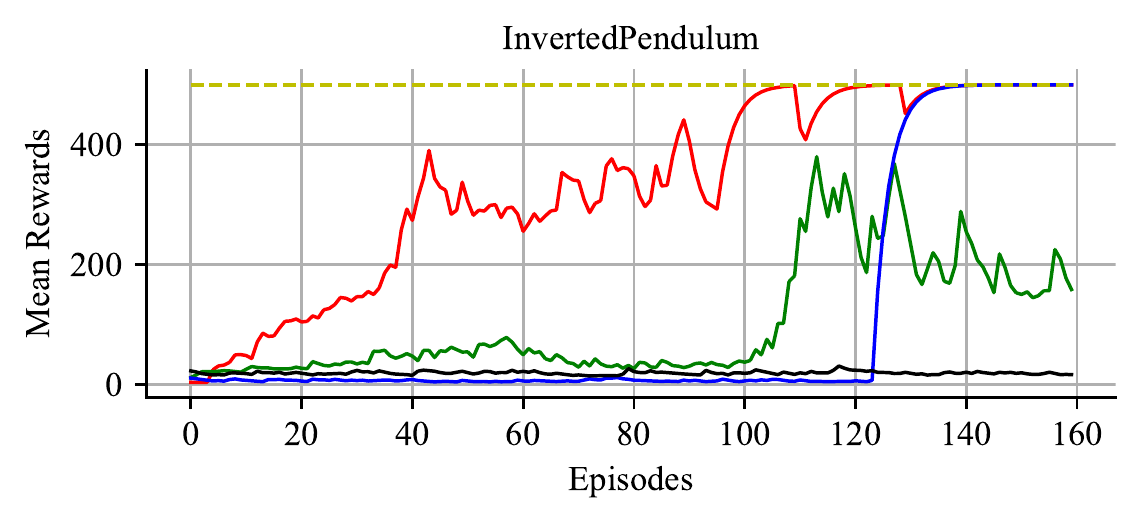}
	}%
\vspace{-5mm}
	\subfigure{
		\centering
		\includegraphics[scale=0.4]{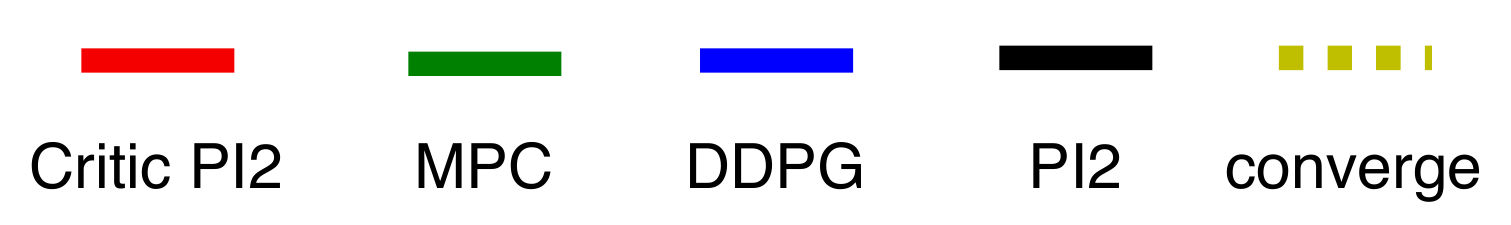}
	}%
	\centering
	\caption{Learning curves of Critic PI2 (ours) and three baselines on different continuous control environments. Each algorithm is evaluated every 1200 environment episodes in InvertedDoublePendulm and 160 episodes in InvertedPendulm, where each evaluation reports the average return over every episode. The actor, critic  and the dynamics are simultaneously trained every two episodes for 100 times \textbf{(For ease of demonstration, DDPG is trained ten times more than other model-based algorithms so that it can converge)}. The dashed reference lines are the asymptotic performance of Critic PI2.}
	\label{fig_Experiments}
\end{figure}

\subsection{Critic PI2 Approach on Benchmark Tasks}

To provide a comparative evaluation against prior methods, we evaluate Critic PI2 on two tasks
(InvertedDoublePendulm, InvertedPendulum) from OpenAI Gym  \cite{gym}, and more details on the experiment platform can be found in Appendix A.
We compare our method to the following state-of-the-art model-based and model-free algorithms:
\begin{enumerate}
    \item \textbf{MPC}  \cite{mpc} This is a widely used model-based reinforcement learning planning algorithm, which represents a comparison to state-of-the-art model-based reinforcement learning. 
    \item \textbf{PI2}  \cite{2016path} Our method builds on top of this model-based algorithm, which makes it a perfect baseline to emphasize the superiority of Critic PI2.
    \item \textbf{DDPG}  \cite{ddpg}  This is an off-policy actor-critic algorithm, which represents a comparison to  state-of-the-art model-free reinforcement learning.
\end{enumerate}
Massive experiments have been carried out on the Mujoco platform  \cite{mujoco}, and relevant results can be found in \textbf{Fig. \ref{fig_Experiments}}, which illustrates the fact that the sample efficiency of
Critic PI2 is far better than both model-based and model-free alternatives. This indicates that
overcoming the representation learning bottleneck, coupled with efficient off-policy Critic, provides for fast learning similar to model-based methods, while attaining final performance comparable to
fully model-free techniques that learn from state. Critic PI2 also substantially outperforms MPC. This
difference can be explained in part by the use of an efficient off-policy Critic, which can
better take advantage of the learned representation and the guarantee of stability obtained by using a trajectory optimization algorithm from stochastic optimal control. 

\begin{figure}[!t]
	\centering
	\subfigure{
		\centering
		\includegraphics[scale=0.7]{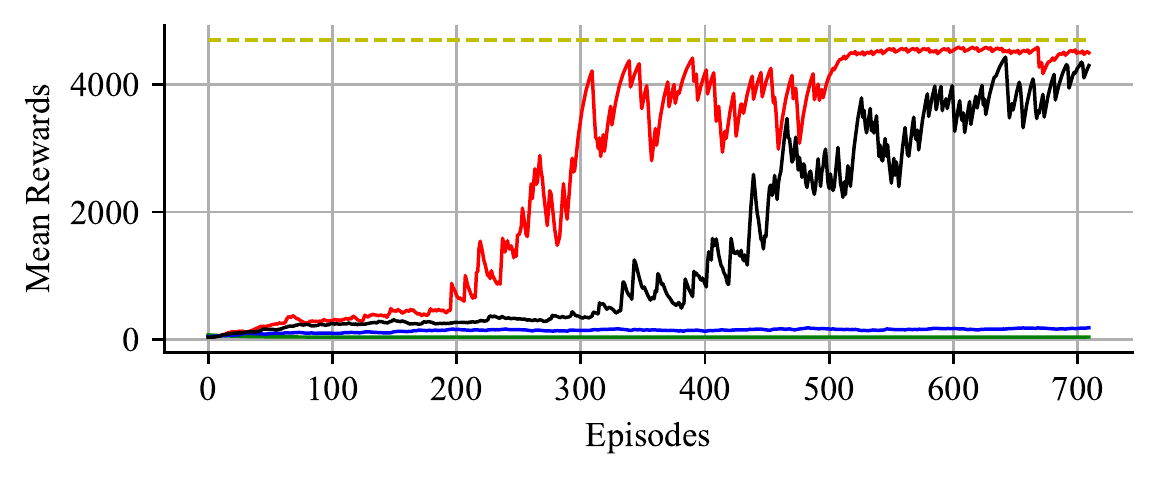}
	}%
	
	\vspace{-5mm}
	\subfigure{
		\centering
		\includegraphics[scale=0.35]{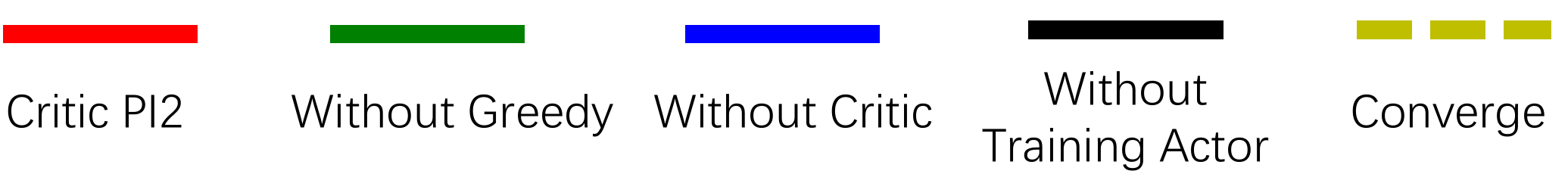}
	}%
	\centering
	\caption{Learning curves of the Ablation Experiment on Critic PI2 (ours), Critic PI2 without greedy strategy while trajectory optimization, Critic PI2 without a Critic to assist the trajectory optimization and planning from  scratch using Critic PI2 without training the actor. Each algorithm is evaluated every 700 environment episodes in InvertedDoublePendulu, where each evaluation reports the average return over every episode. The actor, critic  and the dynamics  are simultaneously trained 100 times  every two episodes. The dashed reference lines are the asymptotic performance of Critic PI2.}
	\label{fig_Ablation Experiment}
\end{figure}

\subsection{Real Time Performance}
As we all know, real-time performance is critical in real-world continues control task, in this section we will investigate Whether Critic PI2 obtain the optimality at the cost of real-time performance.

Different algorithms cost different time while planning an expert policy, and thus affect the real-time performance. ode The predictive horizon $\mathcal{H}$ of MPC and PI2 is 50. We carried out those experiments using a single GTX1660Ti GPU, and an i7-9750H CPU.  The results shown in Table \ref{tab3} prove the fact that Critic PI2 can obtain comparable final performance and  real-time performance  to model-free algorithms like DDPG and 
simultaneously the comparable sample efficiency to  other model-based reinforcement learning algorithm.

\begin{table}[!tp]
	\centering
	\caption{Time Cost for Planning}
	\label{tab3}
	\begin{tabular}{c||c}
		\toprule
		\textbf{Algorithm} &\textbf{Mean Planning Cost (s)}\\
		\midrule
	    Critic PI2 & $0.0139s$  \\
	  PI2 & $1.09s$ \\
	  MPC & $1.22s$ \\
	  DDPG & $0.001s$ \\ 
		\bottomrule
	\end{tabular} 
\end{table}
\subsection{Ablation Experiment}.

We further carry out an ablation experiment to characterize the importance of three main components of our algorithm: 
\begin{enumerate}
    \item\textbf{Without Critic: }No critic to assist the trajectory optimization, just using the Monte Carlo method as \eqref{eq_mc}. 
    \item \textbf{Without Greedy: }Not adopting a greedy strategy while trajectory optimization, but still use the critic.
    \item \textbf{Without Training Actor: }In this settings, the parameters of Actor Network in \textbf{Fig. \ref{fig_actor}} will not be updated during central training.
\end{enumerate}

The results are shown in \textbf{Fig. \ref{fig_Ablation Experiment}}. As we can see, Critic PI2 can obtain a good final performance while ablating the Critic will lead to an extreme performance dropping, which proves the fact that using a critic to assist the predictive model can effectively ease the misleading of the incorrect predictive model, thus guaranteeing good performance. From \textbf{Fig. \ref{fig_Ablation Experiment}}, we can also find out that ablating the greedy strategy will lead to an extreme performance dropping, which demonstrates the fact that only if the planning strategy and the critic work together can the algorithm guarantee a good performance and superior sample efficiency. It's worthy to mention that Critic PI2 algorithm without training the actor can still obtain a good performance at the cost of more sample episodes, which proves the fact that Critic PI2 share 
a reliable training
convergence and optimality of stochastic optimal control and thus can guide the agent to seek out an optimal policy even planning from the scratch.

In conclusions, these experiments can prove the fact that Critic PI2 can plan an expert action even from the scratch and thus perform a new state-of-the-art in both final performance, sample efficiency and real-time performance (compared with other algorithms in Table. \ref{tab3}), whose success can be attributed to both the approximation ability of deep critic network and the dynamic network and the superior planning ability of the PI2 with greedy strategy.

\section{CONCLUSIONS}
\label{phase_conclusion}
In this paper, we present a novel model-based reinforcement learning frameworks, namely Policy Improvement With Path Integrals Using Critic (Critic PI2), simultaneously obtaining the superior learning effect and real-time performance of deep Q learning, great sample efficiency of model-based reinforcement learning and robustness of the stochastic optimal control. Empirical experiment results show asymptotic performance and higher sample
efficiency than previous model-based reinforcement learning algorithms on several benchmark continuous control tasks. For future work, we will investigate the usage of Critic PI2 in other model-based RL frameworks and study how
to leverage frameworks better.




\begin{table*}[tp]
	\centering
	\caption{Environmental Settings}
	\label{tab2}
	\resizebox{1 \linewidth}{!}{
	\begin{tabular}{c||c||c||c||c||c}
		\toprule
		\textbf{Environment Name} &\textbf{Reward Function} &\textbf{Termination Condition}& \textbf{Observation Space Dimension} &\textbf{Action Space Dimension}& \textbf{Steps Per Epoch}\\
		\midrule
	  InvertedPendulum & $1$ if alive & $ \left| \theta_t \right|  \geq 0.2$  & $4$  & $ 1$ & $500$\\
	  InvertedDoublePendulum & $10-0.01x_1^2 - (x_2 - 2)^2 - 10^{-3}\dot{x_1}^2 - 5\times 10^{-3}\dot{x_2}^2$ & $x_2 \leq 1$ & $11$ & $1$ & $500$\\
		\bottomrule
	\end{tabular} 
	}
\end{table*}

\section*{APPENDIX}
\subsection{Experiment Platform}
\label{appendix_a}
\subsubsection{InvertedPendulum} \space This enviornment has a cart sliding on a rail. A pole is connected to the cart. The action is the force applied on the cart along the rail. The actuator force is a real number. The observation includes the angle of the pole away from the upright
vertical position $\theta_t$,  the position of the cart away from the centre of the rail $x_t$ and their first derivative with respect to time (velocity).

\subsubsection{InvertedDoublePendulum} \space This enviornment has a cart sliding on a rail. A pole is connected to the cart while another pole is connected to the pole. The action is the force applied on the cart along the rail. The actuator force is a real number. The observation includes the sine and cosine of the angle of the two poles away from the upright
vertical position $\sin \theta_1$, $\sin \theta_2$, $\cos \theta_1$, $\cos \theta_2$, the position of the cart away from the centre of the rail $x_1$ and the position the top of the second pole away from the centre of the rail $x_2$, their first derivative with respect to time (velocity) $\dot{x_1}, \dot{x_2}, \dot{\theta_1}$, $\dot{\theta_2}$ and constrain force.

 We present our environment settings used in our experiments in Table \ref{tab2}.

\section*{ACKNOWLEDGMENT}
This work is supported  in part by the National Natural Science Foundation of China (62073176), in part by Huawei Noah’s Ark Lab under Grant No.HO2019100802011P313, in part by Natural Science Foundation of Tianjin under Grant 19JCQNJC03200.

\addtolength{\textheight}{-10.3cm}   

\end{document}